\documentclass[a4paper]{article}

\usepackage{INTERSPEECH2015}

\usepackage{graphicx}
\usepackage{amssymb,amsmath,bm}
\usepackage{textcomp}

\ninept

\title{Transferring Knowledge from a RNN to a DNN}

\makeatletter
\def\name#1{\gdef\@name{#1\\}}
\makeatother \name{{\em William Chan$^{*1}$, Nan Rosemary Ke$^{*1}$, Ian Lane$^{1,2}$}}

\address{Carnegie Mellon University \\
  $^1$Electrical and Computer Engineering, $^2$ Language Technologies Institute \\
  $^*$Equal contribution \\
  {\small \tt williamchan@cmu.edu, rosemary.ke@sv.cmu.edu, lane@cmu.edu}
}

\begin{document}

  \maketitle
  
  
  \begin{abstract}
  
  
 Deep Neural Network (DNN) acoustic models have yielded many state-of-the-art results in Automatic Speech Recognition (ASR) tasks. More recently, Recurrent Neural Network (RNN) models have been shown to outperform DNNs counterparts. However, state-of-the-art DNN and RNN models tend to be impractical to deploy on embedded systems with limited computational capacity. Traditionally, the approach for embedded platforms is to either train a small DNN directly, or to train a small DNN that learns the output distribution of a large DNN. In this paper, we utilize a state-of-the-art RNN to transfer knowledge to small DNN. We use the RNN model to generate soft alignments and minimize the Kullback-Leibler divergence against the small DNN. The small DNN trained on the soft RNN alignments achieved a 3.93 WER on the Wall Street Journal (WSJ) eval92 task compared to a baseline 4.54 WER or more than 13\% relative improvement.
 

  \end{abstract}

  \section{Introduction}
 
  Deep Neural Networks (DNNs) combined with Hidden Markov Models (HMMs) have been shown to perform well across many Automatic Speech Recognition (ASR) tasks \cite{dahl-ieeetaslp-2012,deng-icassp-2013,soltau-icassp-2014}. DNNs accept an acoustic context (e.g., a window of fMLLR features) as inputs and models the posterior distribution of the acoustic model. The ``deep'' in DNN is critical, state-of-the-art DNN models often contain multiple layers of  non-linearities, giving it powerful modelling capabilities \cite{zeiler-icassp-2013,dahl-icassp-2013}.

  Recently, Recurrent Neural Networks (RNNs) have demonstrated even more potential over its DNN counterparts \cite{graves-icassp-2013,graves-asru-2013,weng-icassp-2014}. RNN models are neural network models that contain recurrent connections or cycles in the connectivity graph. RNN models when unrolled, can actually be seen as a very special case of DNN. The recurrent nature of the RNN allows us to model temporal dependencies, which is often the case in speech sequences. In particular, the recurrent structure of the model allows us to store temporal information (e.g., the cell state in LSTM \cite{hochreiter-neuralcomputation-1997}) within the model. In \cite{sak-interspeech-2014}, RNNs were shown to outperform DNNs in large commercial ASR systems. And in \cite{weng-icassp-2014}, RNNs have been shown to provide better performance over DNNs in robust ASR.
  
  Currently, there has been much industry interest in ASR for embedded platforms, for example, mobile phones, tablets and smart watches. However, these platforms tend to have limited computational capacity (e.g., no/limited GPU and/or low performance CPU), limited power availability (e.g., small batteries) and latency requirements (e.g., asking a GPS system for driving directions should be responsive). Unfortunately, many state-of-the-art DNN and RNN models are simply too expensive or impractical to run on embedded platforms. Traditionally, the approach is simply to use a small DNN, reducing the number of layers and the number of neurons per layer; however, such approaches often suffer from Word Error Rate (WER) performance degradations \cite{lei-interspeech-2013}. In our paper, we seek to improve the WER of small models which can be applied to embedded platforms.
  
  DNNs and RNNs are typically trained from forced alignments generated from a GMM-HMM system. We refer to this as a hard alignment, the posterior distribution is concentrated on a single acoustic state for each acoustic context. There has been evidence that these GMM alignment labels are not the optimal training labels as seen in \cite{navdeep-interspeech-2014,senior-icassp-2014}. The GMM alignments make various assumptions of the data, such as independence of acoustic frames given states \cite{navdeep-interspeech-2014}. In this paper, we show soft distribution labels generated from an expert is potentially more informative over the GMM hard alignments leading to WER improvements. The effects of the poor GMM alignment quality may be hidden away in large deep networks, which have sufficient model capacity. However, in narrow shallow networks, training with the same GMM alignments often hurts our ASR performance \cite{lei-interspeech-2013}.
  
  One approach is to change the training criteria, rather than trying to match our DNN to the GMM alignments, we can instead try and match our DNN to the distribution of an expert model (e.g., a big DNN). In \cite{li-interspeech-2014}, a small DNN was trained to match the output distribution of a large DNN. The training data labels are generated by passing labelled and unlabelled data through the large DNN, and training the small DNN to match the output distribution. The results were promising, \cite{li-interspeech-2014} achieved a 1.33\% WER reduction over their baseline systems.
  
  Another approach is to train an model to match the softmax logits of an expert model. In \cite{hinton-nips-2014}, an ensemble of experts were trained and used to teach a (potentially smaller) DNN. Their motivation was inference (e.g., computational cost grows linearly to the number of ensemble models), however the principle of model compression applies \cite{bucila-sigkidd-2006}. \cite{hinton-nips-2014} also generalized the framework, and showed that we can train the models to match the logits of the softmax, rather than directly modelling the distributions which could yield more knowledge transfer.

  In this paper, we want to maximize small DNN model performance targeted at embedded platforms. We transfer knowledge from a RNN expert to a small DNN. We first build a large RNN acoustic model, and we then let the small DNN model learn the distribution or soft alignment from the large RNN model. We show our technique will yield improvements in WER compared to the baseline models trained on the hard GMM alignments.
  
  The paper is structured as follows. Section 2, begins with an introduction of a  state-of-the-art RNN acoustic model. In Section 3, we describe the methodology used to transfer knowledge from a large RNN model to a small DNN model. Section 4 is gives experiments, results and analysis. And we finish in Section 5 with our conclusion and future work discussions.

  \section{Deep Recurrent Neural Networks}
    There exist many implementations of RNNs \cite{chung-nips-2014}, and LSTM is a particular implementation of RNN that is easy to train and does not suffer from the vanishing or exploding gradient problem in Backpropagation Through Time (BPTT) \cite{hochreiter-2011}. We follow \cite{vinyals-arvix-2014,chan-interspeech-2015} in our LSTM implementation:
  \begin{align}
      i_t &= \phi(W_{xi} x_t + W_{hi} h_{t - 1}) \\
      f_t &= \phi(W_{xf} x_t + W_{hf} h_{t - 1}) \\
      c_t &= f_t \odot cs_{t - 1} + i_t \odot \tanh(W_{xc} x_t + W_{hc} h_{t - 1}) \\
      o_t &= \phi(W_{xo} x_t + W_{ho} h_{t - 1}) \\
      h_t &= o_t \odot \tanh(c_t)
    \end{align}
  This particular LSTM implementation omits the the bias and peephole connections. We also apply a cell clipping of 3 to ease the optimization to avoid exploding gradients. LSTMs can also be extended to be a Bidirectional LSTM (BLSTM), to capture temporal dependencies in both set of directions \cite{graves-asru-2013}.
  
  RNNs (and LSTMs) can be also be extended into deep RNN architectures \cite{pascanu-iclr-2014}. There has been evidence that the deep RNN models can perform better than the shallow RNN models \cite{graves-asru-2013,pascanu-iclr-2014,chan-interspeech-2015}. The additional layers of nonlinearities can give the network additional model capacity similar to the multiple layers of nonlinearities in a DNN.
  
  We follow \cite{chan-interspeech-2015}, in building our deep RNN; to be exact, the particular RNN model is actually termed a TC-DNN-BLSTM-DNN model. The architecture begins with a Time Convolution (TC) over the input features (e.g., fMLLR) \cite{chan-icassp-2015}. This is followed by a DNN signal processor which can project the features into a higher dimensional space. The projected features are then consumed by a BLSTM, modelling the acoustic context sequence. Finally a DNN with a softmax layer is used to model the posterior distribution. \cite{chan-interspeech-2015}'s model gave more than 8\% relative improvement over previous state-of-the-art DNNs in the Wall Street Journal (WSJ) eval92 task. In this paper, we use the TC-DNN-BLSTM-DNN model as our deep RNN to generate the training alignments from which the small DNN will learn from.


  \section{Methodology}
  Our goal is to transfer knowledge from the RNN expert to a small DNN. We follow an approach similar to \cite{li-interspeech-2014}. We transfer knowledge by training the DNN to
match the RNN's output distribution. Note that we train on the soft distribution of the RNN (e.g., top $k$ states) rather than just the top-1 state (e.g., realigning the model with the RNN). In this paper we will show the distribution generated by the RNN is more informative over the GMM alignments. We will also show the soft distribution of the RNN is more informative over taking just the top-1 state generated by the RNN.

  \subsection{KL Divergence}
  
  We can match the output distribution of our DNN to our RNN by minimizing the Kullback-Leibler (KL) divergence between the two distributions. Namely, given the RNN posterior distribution $P$ and the DNN posterior distribution $Q$, we want to minimize the KL divergence $D_{KL}(P || Q)$:
  \begin{align}
  D_{KL}(P(s | x) || Q(s | x)) &= \sum_i P(s_i | x) \ln \frac{P(s_i | x)}{Q(s_i | x)} \\
                                 &= H(P, Q) - H(P)
  \end{align}
  where $s_i \in s$ are the acoustic states, $H(P, Q) = \sum_i -P(s_i | x) \ln Q(s_i | x)$ is the cross entropy term and $H(P) = \sum_i P(s_i | x) \ln P(s_i | x)$ is the entropy term. We can safely ignore the $H(P)$ entropy term since its gradient is zero with respect to the small DNN parameters. Thus, minimizing the KL divergence is equivalent to minimizing the Cross Entropy Error (CSE) between the two distributions:
  \begin{align}
     H(P, Q) = \sum_i -P(s_i | x) \ln Q(s_i | x)
  \end{align}
  which we can easily differentiate and compute the pre-softmax activation $a$ (e.g., the softmax logits) derivative:
  \begin{align}
  \frac{\partial J}{\partial a_i} = Q(s_ i | x) - P(s_i | x)
  \end{align}
  
  \begin{figure*}[t]
  \centering
  \includegraphics[width=0.75\linewidth]{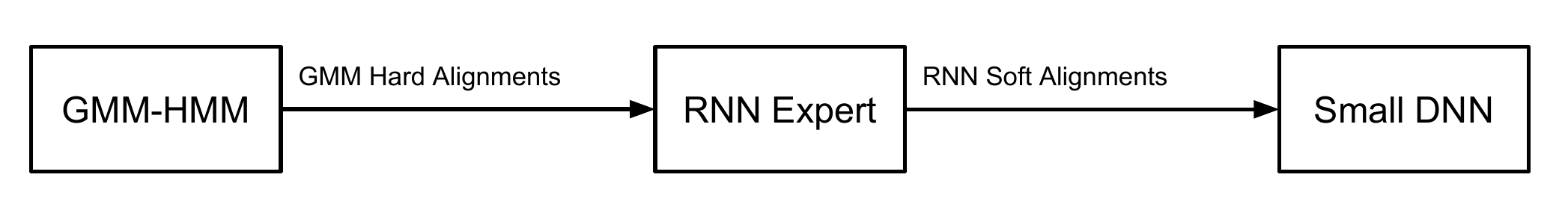}
  \caption{We use the hard GMM alignments to first train a RNN, after which we use the soft alignments from the RNN to train our small DNN.}
  \label{fig:workflow}
  \end{figure*}
  
  \subsection{Alignments}
  
  In most ASR scenarios, DNNs and RNNs are typically trained with forced alignments generated from GMM-HMM models to model the posterior distribution. We refer this alignment as a hard GMM alignment because the probability is concentrated on only a single state. Furthermore, the alignment labels generated from GMM-HMM model are not always the optimal for training DNNs \cite{navdeep-interspeech-2014}. The GMM-HMM makes various assumptions that may not be true (e.g., independence of frames). One possible solution is to use labels or alignments from another expert model, for example in \cite{hinton-nips-2014} an ensemble of experts was used to teach one model. In this paper, we generate labels from an expert RNN which provide better training targets compared to the GMM alignments.
  
  One possibility is to generate hard alignments from a RNN expert. This is done by first training the RNN with hard alignments from the GMM-HMM model. After the DNN is trained, we then realign the data by taking hard alignments (e.g., top-1 probability state) from the trained RNN. The alignment is hard as it takes only the most probable phoneme state for each acoustic context, and the probability is concentrated on a single phoneme state. 
  
  On the other hand, we could utilize the full distribution or soft alignment associated with each acoustic frame. More precisely, for each acoustic context, we take the full distribution of the phonetic states and their probabilities. However, this suffers from several problems. First, during training, we need to either run the RNN in parallel or pre-cache the distribution on disk. Running the RNN in parallel is an expensive operation and undesirable. The alternative is caching the distribution on disk, which would require obscene amounts of storage (e.g., we typically have several thousand acoustic states). For example, in WSJ, it would take over 30 TiB to store the full distribution of the si284 dataset. We also run into bandwidth issues when loading the training samples from the disk cache. Finally, the entire distribution may not be useful, as there will be many states with near zero values; intuition suggests we can just discard those states (e.g., lossy compression).
  
  Our solution sits inbetween the two extremes of taking only the top-1 state or taking the full distribution. We find that the posterior distributions are typically concentrated on only a few states. Therefore, we can make use of almost the full distribution by storing only a small portion of the states probability distribution. We take the states that contains the top 98\% of the probability distribution. Note, this is different than taking the top-$k$ states, we take at least $n$ states where we can capture at least 98\% of the distribution, and $n$ will vary per frame. We then re-normalize the probability per frame to ensure the distribution sums up to 1. This lossy compression method losses up to 2\% of the original probability mass.
  

  
  
  
  
  
  
 
  
  \section{Experiments and Results}
   We experiment with the WSJ dataset; we use si284 with approximately 81 hours of speech as the training set, dev93 as our development set and eval92 as our test set. We observe the WER of our development set after every epoch, we stop training once the development set no longer improves. We report the converged dev93 and the corresponding eval92 WERs. We use the same fMLLR features generated from the Kaldi s5 recipe \cite{povey-asru-2011}, and our decoding setup is exactly the same as the s5 recipe (e.g., big dictionary and trigram pruned language model). We use the tri4b GMM alignments as our hard forced alignment training targets, and there are a total of 3431 acoustic states. The GMM tri4b baseline achieved a dev and test WER of 9.39 and 5.39 respectively.
   
   \subsection{Optimization}
   
   In our DNN and RNN optimization procedure, we initialized our networks randomly (e.g., no pretraining) and we used Stochastic Gradient Descent (SGD) with a minibatch size of 128. We apply no gradient clipping or gradient projection in our LSTM. We experimented with constant learning rates of [0.1, 0.01, 0.001] and geometric decayed learning rates with initial values of [0.1, 0.01] with a decay factor of 0.5. We report the best WERs out of these learning rate hyperparameter optimizations.

  \subsection{Big DNN and RNN}

  We first built several baseline (big) DNN and RNN systems. These are the large networks and not suitable for deployment on mobile platforms. We followed the Kaldi s5 recipe and built a 7 layer DNN and 2048 neurons per hidden layer with DBN pretraining and achieves a eval92 WER of 3.81 \cite{povey-asru-2011}. We also followed \cite{chan-interspeech-2015} and built a 5 layer ReLU DNN with 2048 neurons per hidden layer and achieves a eval92 WER of 3.79. Our RNN model follows \cite{chan-interspeech-2015}, consists of 2048 neurons per layer for the DNN layers, and 256 bidirectional cells for the BLSTM. The RNN model achieves a eval92 WER of 3.47, significantly better than both big DNN models. Each network has a softmax output of 3431 states matching the GMM model. Table \ref{tab:dnn} summarizes the results for our baseline big DNN and big RNN experiments.

  \begin{table}[t]
\caption{\label{tab:dnn} {\it Wall Street Journal WERs for big DNN and RNN models.}}
\vspace{2mm}
\centerline{
\begin{tabular}{|c|c|c|}
\hline
\bfseries Model & dev93 WER & eval92 WER \\
\hline
\hline
GMM Kaldi          & 9.39 & 5.39 \\
DNN Kaldi s5       & 6.68 & 3.81 \\
DNN ReLU           & 6.84 & 3.79 \\
RNN \cite{chan-interspeech-2015} & 6.58 & 3.47 \\
\hline
\end{tabular}
}
\end{table}

  \subsection{Small DNN}

  We want to build a small DNN that is easily computable by an embedded device. We decided on a 3 layer network (2 hidden layers), wherein each hidden layer has 512 ReLU neurons and a final softmax of 3431 acoustic states matching the GMM. Since Matrix-Matrix Multiplication (MMM) is an $O(n^3)$ operation, the effect is approximately a 128 times reduction in number of computations for the hidden layers (when comparing the 4 hidden layers of 2048 neurons vs. a 2 hidden layers of 512 neurons). This will allow us to perform fast interference on embedded platforms with limited CPU/GPU capacity.

  We first trained a small ReLU DNN using the hard GMM alignments. We achieved a 4.54 WER compared to 3.79 WER of the big ReLU DNN model on the eval92 task. The dev93 WER is 8.00 for small model vs 6.84 for the large model; the big gap in dev93 WER suggests the big DNN model is able to optimize substantially better. The large DNN model has significantly more model capacity, and thus yielding its better results over the small DNN.
  
  Next, we experimented with the hard RNN alignment. We take the top-1 state of the RNN model and train our DNN towards this alignment. We did not see any improvement, while the dev93 WER improves from 8.00 to 7.83, the eval92 WER degrades from the 4.54 to 4.63. This suggests, the RNN hard alignments are worse labels than the original GMM alignments. The information provided by the RNN when looking at only the top state is no more informative over the GMM hard alignments. One hypothesis is our DNN model overfits towards the RNN hard alignments, since the dev93 WER was able to improve, while the model is unable to generalize the performance to the eval92 test set.
  
  We now experiment with the RNN soft alignment, wherein we can add the soft distribution characteristics of the RNN to the small DNN. We take the top 98\% percentile of probabilities of from the RNN distribution and renormalize them (e.g., ensure the distribution sums up to $1$). We minimize the KL divergence between the RNN soft alignments and the small DNN. We see a significant improvement in WER. We achieve a dev93 WER of 7.38 and eval92 3.93. In the eval92 scenario, our WER improves by over 13\% relative compared to the baseline GMM hard alignment. We were almost able to match the WER of the big DNN of 3.79 (off by 3.6\% relative), despite the big DNN have many more layers and neurons. The RNN soft alignment adds considerable information to the training labels over the GMM hard alignments or the RNN hard alignments.
  
  We also experimented training on the big DNN soft alignments. The big DNN model is the DNN ReLU model mentioned in table \ref{tab:dnn}, wherein it achieved a eval92 WER of 3.79. Once again, we generate the soft alignments and train our small DNN to minimize the KL divergence. We achieved a dev93 WER of 7.43 and eval92 WER of 4.27. There are several things to note, first, we once again improve over the GMM baseline by $5.9$\% relative. Next, the dev93 WER is very close to the RNN soft alignment (less than $1$\% relative), however, the gap widens when we look at the eval92 WER (more than $8$\% relative). This suggests the model overfits more under the big DNN soft alignments, and the RNN soft alignments provide more generalization. The quality of the RNN soft alignments are much better than big DNN soft alignments. Table \ref{tab:smalldnn} summarizes the WERs for the small DNN model using different training alignments.

\begin{table}[t]
\caption{\label{tab:smalldnn} {\it Small DNN WERs for Wall Street Journal based on different training alignments.}}
\vspace{2mm}
\centerline{
\begin{tabular}{|c|c|c|}
\hline
\bfseries Alignment & dev93 WER & eval92 WER \\
\hline
\hline
Hard GMM & 8.00 & 4.54 \\
Hard RNN & 7.83 & 4.63 \\
Soft RNN & 7.38 & 3.93 \\
Soft DNN & 7.43 & 4.27 \\
\hline
\end{tabular}
}
\end{table}

\subsection{Cross Entropy Error}
We compute the CSE of our various models against the GMM alignment for the dev93 dataset. We measure the CSE against dev93 since that is our stopping criteria and that is the optimization loss. The CSE will give us a better indication of the optimization procedure, and how our models are overfitting. Table \ref{tab:cse} summarizes our CSE measurements. There are several observations, first the big RNN is able to achieve a lower CSE compared to the big DNN. The RNN model is able to optimize better than the DNN as seen with the better WERs the RNN model provides. This is as expected since the big RNN model achieves the best WER.

The next observation is that the small DNNs trained off the soft alignment from the large DNN or RNN achieved a lower CSE and compared to the small DNN trained on the GMM hard alignment. This suggests the soft alignment labels are indeed better training labels in optimizing the model. The extra information contained in the soft alignment helps us optimize better towards our dev93 dataset.

The small DNN trained on the soft RNN alignments and soft DNN alignments give interesting results. These models achieved a lower CSE compared to the large RNN and large DNN models trained on the GMM alignments. However, the WERs are worse than the large RNN and large DNN models. This suggests the small model trained on the soft distribution is overfitting, it is unclear if the overfitting occurs because the smaller model can not generalize as well as the large model, or if the overfitting occurs because of the quality of the soft alignment labels.




\begin{table}[t]
\caption{\label{tab:cse} {\it Cross Entropy Error (CSE) on WSJ dev93 over our various models.}}
\vspace{2mm}
\centerline{
\begin{tabular}{|c|c|c|c|}
\hline
\bfseries Alignment & Model &  CSE  \\
\hline
\hline
GMM & Big RNN & 1.27620  \\
GMM & Big DNN & 1.28431  \\
\hline
Hard RNN & Small DNN & 1.52135 \\
Soft RNN & Small DNN & 1.24617 \\
Soft DNN & Small DNN & 1.24723 \\
\hline
\end{tabular}
}
\end{table}

  \section{Conclusion and Discussions}
  The motivation and application of our work is to extend ASR onto embedded platforms, where there is limited computational capacity. In this paper we have introduced a method to transfer knowledge from a RNN to a small DNN. We minimize the KL divergence between the two distributions to match the DNN's output to the RNN's output. We improve the WER from 4.54 trained on GMM forced alignments to 3.93 on the soft alignments generated by the RNN. Our method has resulted in more than 13\% relative improvement in WER with no additional inference cost.
  
  One question we did not answer in this paper is whether the small DNN's model capacity or the RNN's soft alignment is the bottleneck of further WER performance. We did not measure the effect of the small DNN's model capacity on the WER, would we get similar WERs if we increased or decreased the small DNN's size? If the bottleneck is in the quality of the soft alignments, then in princple we could reduce the small DNN's size further without impacting WER (much), however, if model capacity is the issue, then we should not use smaller networks.
  
  On a similar question, we did not investigate the impact of the top probability selection in the RNN alignment. We threshold the top 98\% of the probabilities out of convenience, however, how would selecting more or less probabilities affect the quality of the alignments. In the extreme case, wherein we only selected the top-1 probability, we found the model to perform much worse compared to the 98\% soft alignments, and even worse than the GMM alignments, this evidence definitely shows the importance of the information contained in the soft alignment.
  
  We could also extend our work similar to \cite{li-interspeech-2014} and utilize vast amounts of unlabelled data to improve our small DNN. In \cite{li-interspeech-2014}, they applied unlabelled data to their large DNN expert to generate vast quantities of soft alignment labels for the small DNN to learn from. In principle, one could extend this to an infinite amount of training data with synthetic data generation, which has been shown to improve ASR performance \cite{awni-arxiv-2014}.
  
  Finally, we did not experiment with sequence training \cite{vesely-interspeech-2013}, sequence training has almost always shown to help \cite{sak-interspeech-2014b}, it would be interesting to see the effects of sequence training on these small models, and whether we can further improve the ASR performance.
  
  

  \section{Acknowledgements}
  We thank Won Kyum Lee for helpful discussions and proofreading this paper.

    \newpage

  \eightpt
  \bibliographystyle{IEEEtran}
  \bibliography{paper}
\end{document}